\documentclass[sigconf, screen]{acmart}%%
\usepackage{subcaption}
\usepackage{multirow}
\usepackage{balance}
\AtBeginDocument{%
  }

\copyrightyear{2025}
\acmYear{2025}
\setcopyright{cc}
\setcctype{by}
\acmConference[DFF '25]{Proceedings of the 1st Deepfake Forensics Workshop:
Detection, Attribution, Recognition, and Adversarial Challenges in the Era of
AI-Generated Media}{October 27--28, 2025}{Dublin, Ireland}
\acmBooktitle{Proceedings of the 1st Deepfake Forensics Workshop: Detection,
Attribution, Recognition, and Adversarial Challenges in the Era of AI-Generated
Media (DFF '25), October 27--28, 2025, Dublin,
Ireland}\acmDOI{10.1145/3746265.3759670}
\acmISBN{979-8-4007-2047-5/2025/10}

\begin{document}

\title[Bridging the Gap]{Bridging the Gap: A Framework for Real-World Video Deepfake Detection via Social Network Compression Emulation}

\author{Andrea Montibeller}
\email{andrea@truebees.eu}
\orcid{0000-0002-0794-312X}
\affiliation{%
  \institution{University of Trento and\\ Truebees srl}
  \city{Trento}
  \country{Italy}
}

\author{Dasara Shullani}
\email{dasara.shullani@unifi.it}
\orcid{0000-0003-2753-366X}
\affiliation{%
  \institution{University of Florence}
  \city{Firenze}
  \country{Italy}
  }

\author{Daniele Baracchi}
\email{daniele.baracchi@unifi.it}
\orcid{0000-0002-7364-1955}
\affiliation{%
  \institution{University of Florence}
  \city{Firenze}
  \country{Italy}
  }

\author{Alessandro Piva}
\email{alessandro.piva@unifi.it}
\orcid{0000-0002-3047-0519}
\affiliation{%
  \institution{University of Florence}
  \city{Firenze}
  \country{Italy}
  }

\author{Giulia Boato}
\email{giulia.boato@truebees.eu}
\orcid{0000-0002-0260-9528}
\affiliation{%
  \institution{University of Trento\\ and Truebees srl}
  \city{Trento}
  \country{Italy}
}

\begin{abstract}
The growing presence of AI-generated videos on social networks poses new challenges for deepfake detection, as detectors trained under controlled conditions often fail to generalize to real-world scenarios. A key factor behind this gap is the aggressive, proprietary compression applied by platforms like YouTube and Facebook, which launder low-level forensic cues. However, replicating these transformations at scale is difficult due to API limitations and data-sharing constraints.
For these reasons, we propose a first framework that emulates the video sharing pipelines of social networks by estimating compression and resizing parameters from a small set of uploaded videos. These parameters enable a local emulator capable of reproducing platform-specific artifacts on large datasets without direct API access.
Experiments on FaceForensics++ videos shared via social networks demonstrate that our emulated data closely matches the degradation patterns of real uploads. Furthermore, detectors fine-tuned on emulated videos achieve comparable performance to those trained on actual shared media.
Our approach offers a scalable and practical solution for bridging the gap between lab-based training and real-world deployment of deepfake detectors, particularly in the underexplored domain of compressed video content.
\end{abstract}

\begin{CCSXML}
<ccs2012>
   <concept>
       <concept_id>10010147.10010178</concept_id>
       <concept_desc>Computing methodologies~Artificial intelligence</concept_desc>
       <concept_significance>500</concept_significance>
       </concept>
   <concept>
       <concept_id>10010405.10010462.10010464</concept_id>
       <concept_desc>Applied computing~Investigation techniques</concept_desc>
       <concept_significance>500</concept_significance>
       </concept>
   <concept>
       <concept_id>10010147.10010178.10010224</concept_id>
       <concept_desc>Computing methodologies~Computer vision</concept_desc>
       <concept_significance>300</concept_significance>
       </concept>
 </ccs2012>
\end{CCSXML}

\ccsdesc[500]{Computing methodologies~Artificial intelligence}
\ccsdesc[500]{Applied computing~Investigation techniques}
\ccsdesc[300]{Computing methodologies~Computer vision}

\keywords{Multimedia forensics, AI-generated video detection, Deepfakes, Social Networks}


\maketitle

\section{Introduction}
The recent advancements in generative AI have led to the widespread democratization and proliferation of AI-generated media.
Derivations of techniques such as Stable Diffusion, FLUX.1, DeepFloyd IF, Kandinsky, VQGAN+CLIP, and AnimateDiff
\cite{rombach_high-resolution_2022, podell2023sdxl, flux2024, deepfloyd2023if, kandinsky2023, esser2021taming, animatediff2023}, which once required significant technical expertise and a setup involving downloads from platforms like GitHub or Hugging Face, are now integrated into user-friendly web applications like ChatGPT and Gemini or made directly available through commercial software such as Adobe Creative Suite. Furthermore, generative AI techniques are not restricted to image synthesis alone; they extend to audio \cite{audio_DM} and video \cite{battocchio2025advance} as well, enabling a wide range of global and local generative editing tasks, including audio and video reenactment, face swapping \cite{thies2016face2face}, voice spoofing \cite{todisco2019asvspoof}, and inpainting~\cite{bertazzini2024beyond}.

While AI-generated media are, at the moment, predominantly employed in entertainment and advertising \cite{amerini2024deepfake, amerini2025deepfake}, recent years have seen an alarming trend of their use on social media platforms to disseminate fake news and misleading or harmful content \cite{dell2025truefake}.

To protect citizens from falling victim to such deceptive media, the multimedia forensics research community has developed various deepfake detection techniques. In the domain of fake image detection, convolutional neural network (CNN)-based models such as ResNet50, DenseNet, EfficientNet, and XceptionNet \cite{koonce2021resnet, arulananth2024classification, koonce2021efficientnet, lu2022deep} have been widely used to analyze low-level features and detect subtle inconsistencies introduced during the generative process. Similar approaches leveraging large pre-trained networks \cite{battocchio2025advance, cozzolino_raising_2024, wang2023low}, such as Vision Transformer (ViT), have also recently been proposed for detecting fake images, videos and audio, achieving excellent performance in terms of accuracy and generalization when tested under controlled laboratory conditions.

Nevertheless, state-of-the-art (SoA) methods developed and evaluated under controlled laboratory conditions, often struggle when applied to real-world scenarios such as media sharing on social networks \cite{dell2025truefake, boato_trueface_2022, marcon2021detection}. While these real-world scenarios are inherently complex to address during development (often involving cumbersome, costly, and rate-limited APIs)\footnote{\url{https://docs.x.com/x-api/getting-started/about-x-api}}, neglecting them has led to substantial deepfake detectors performance degradation and reduced accuracy \cite{8397040, dell2025truefake, marcon2021detection, boato_trueface_2022}.
Moreover, prior works has primarily focused on fake image detection, benefiting from easier access to social media image uploads via APIs, while the video domain remained significantly underexplored \cite{marcon2021detection, dell2025truefake}.

For these reasons, we propose a novel framework specifically designed to model social network video processing for deepfake detection in the real-world.
Our framework requires fewer than 50 uploaded videos per resolution to a target social network in order to infer platforms crucial re-encoding parameters, used for compression and resizing. These parameters are then stored in a local database and applied to any dataset of non-shared videos to accurately emulate the social network’s video processing. %This enables more realistic training and evaluation of deepfake detection models, without the overhead of direct API interaction, such as high costs, rate limits, or sharing quotas. 
The proposed approach enables realistic model training and evaluation while bypassing API-related constraints like high costs, rate limits, and content-sharing quotas.

Experimental results, obtained using both legacy videos from the FaceForensics++ (FF++) dataset \cite{marcon2021detection} and newer videos we shared across major platforms in 2025, demonstrate high modeling accuracy on content from Facebook, YouTube, and BlueSky, spanning several years.

The rest of this paper is organized as follows: Section~\ref{sec:SoA} discusses the impact of social network media compression on deepfake detectors. Section~\ref{sec:Method} presents the proposed emulation framework. Experimental results and ablation studies are detailed in Section~\ref{sec:Exp}, and conclusions are provided in Section~\ref{sec:Conc}.

\section{Related Work}
\label{sec:SoA}

A particularly challenging scenario in multimedia forensics is the detection of deepfake content shared on social networks~\cite{amerini2024deepfake, amerini2025deepfake,dell2025truefake,8397040,pasquini2021media}. To address bandwidth and storage constraints~\cite{boato_trueface_2022, dell2025truefake}, social media platforms often apply aggressive compression and resizing techniques. While these measures reduce file sizes, they also significantly degrade forensic features that are essential for distinguishing authentic content from manipulated media~\cite{boato_trueface_2022,dell2025truefake,Verdoliva2020910,maier2024reliable}.

In an effort to curb the proliferation of fake profile images on social platforms, Yang et al.~\cite{yang2024characteristics} conducted a systematic study of such accounts on X (formerly Twitter). They introduced a heuristic based on eye alignment patterns in GAN-generated faces and discovered that these images often share consistent eye coordinates. This insight enabled them to estimate the presence of approximately 10{,}000 active accounts per day using synthetic faces.

Similarly, Ricker et al.~\cite{ricker2024ai} performed a large-scale analysis of AI-generated profile images on X. They proposed a multi-stage detection pipeline based on ResNet50~\cite{koonce2021resnet}, corroborating Yang et al.'s estimates regarding the prevalence of GAN-generated profiles.

Extending the research beyond GAN-based imagery, Maier and Riess~\cite{maier2024reliable} investigated the detection of synthetic images generated by diffusion models. They proposed a Bayesian Neural Network (BNN) capable of delivering reliable uncertainty estimates, particularly effective in identifying out-of-distribution samples while maintaining competitive accuracy with state-of-the-art detectors. In contrast, Kumar et al.~\cite{kumar2023gan} focused on GAN-based convolutional networks for both the generation and detection of deepfakes, offering a comparative analysis using metrics such as the Inception Score (IS) and Fréchet Inception Distance (FID).

To bridge the gap between controlled experimental conditions and real-world applications, Boato et al.~\cite{boato_trueface_2022} introduced a large and diverse dataset containing 80{,}000 fake images generated with StyleGAN and 70{,}000 real images sourced from leading benchmark datasets~\cite{karras_style-based_2019}. Their study assessed the compression levels applied by platforms such as Twitter, Facebook, and Telegram, enabling the fine-tuning of detection models without incurring catastrophic forgetting~\cite{french1999catastrophic}.

Building on this work, Dell'Anna et al.~\cite{dell2025truefake} introduced \textit{TrueFake}, an updated version of the previous dataset. It includes 600{,}000 images generated using the latest GAN and diffusion models, along with real images captured using modern camera pipelines with AI-based image enhancement. A significant subset of \textit{TrueFake} was distributed across three major social networks (Facebook, X, and Telegram) to analyze degradation effects. Evaluation metrics such as Peak Signal-to-Noise Ratio (PSNR) and Structural Similarity Index Measure (SSIM) were used to assess the impact of compression, supporting the fine-tuning and benchmarking of SoA detection models under realistic conditions.

While most deepfake detection research has concentrated on still images, the detection of manipulated videos on platforms like YouTube and Facebook remains underexplored. This gap is largely due to technical challenges such as API limitations, large file sizes, and high processing demands. The limited existing studies~\cite{marcon2021detection, hu2021detecting} underscore the importance of video deepfake detection but also highlight obstacles such as platform-specific compression, resolution variability, and data-sharing constraints. Moreover, the constant evolution of social media processing pipelines and rapid advancements in generative models necessitate ongoing updates to datasets and evaluation protocols, making the collection and maintenance of large-scale, real-world video datasets particularly difficult.

To address these limitations, the following sections present a novel framework designed to emulate social network video compression using a predefined set of shared videos. This approach enables the evaluation and fine-tuning of SoA detectors on realistically degraded media without relying on social media APIs, thereby avoiding associated costs and usage limitations.

\section{Social Network Video Sharing Emulator}
\label{sec:Method}

\begin{figure}[!t]
    \centering
    \includegraphics[width=\linewidth]{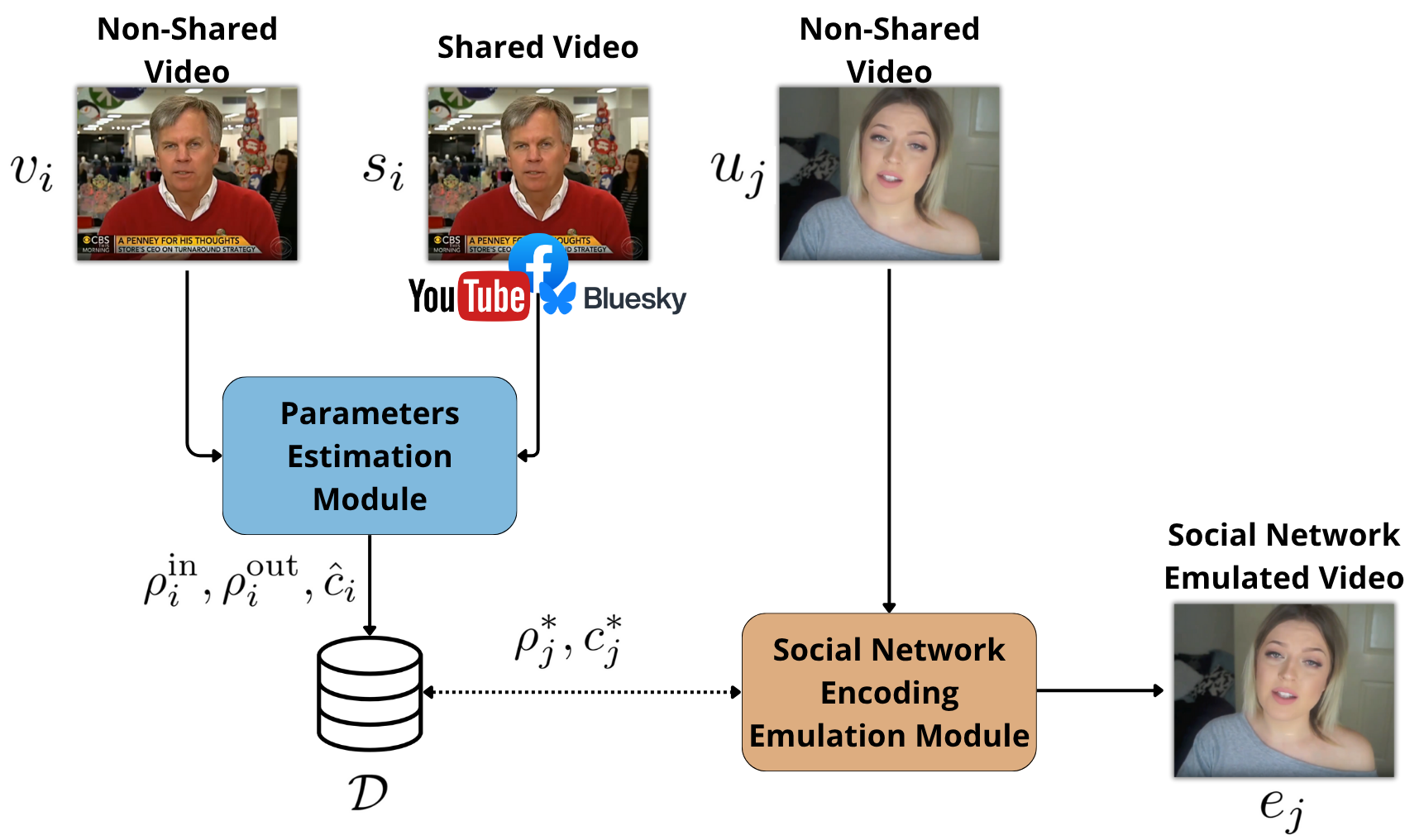}
    \caption{Overview of the Social Network Video Sharing Emulator framework. The system consists of a Parameters Estimation Module, which estimate compression and resolution parameters from pairs of original $v_i$ and social-network-shared videos $s_i$; and the Social Network Encoding Emulation Module that use the estimated parameters to emulate social networks compression effects on new videos $u_j$.
}
    \label{fig:framework}
\end{figure}

The proposed Social Network Video Sharing Emulator (SNVSE)\footnote{Full-code is available at \url{https://github.com/truebees-ai/social_emulator}.}, visible in Fig.~\ref{fig:framework}, is designed to replicate the compression artifacts applied by popular social media platforms (e.g. {YouTube} and {Facebook}) to video content. To do so, a small set of $N$ videos $\mathcal{V}$ have to be uploaded on the social network we want to emulate producing a new set of $N$ social network shared videos $\mathcal{S}$. Thus, two separate modules of the proposed SNVSE: the \textit{Parameters Estimation Module} and the \textit{Social Network Encoding Emulation Module}; operate in succession, forming a pipeline that first estimates the effective compression rate factor (CRF) and the respective resizing parameter applied by the social network, and then re-encodes non-shared videos to match this compression behavior. In the following subsections we provide details on each module.

\subsection{Parameters Estimation Module}
\label{ssec:crf}

Let $\mathcal{V} = \{v_1, v_2, \dots, v_N\}$, denote a set of $N$ reference pre-social network upload videos of a given resolution, and $\mathcal{S} = \{s_1, s_2, \dots, s_N\}$ denote the corresponding set of social network-shared videos. For each video $v_i \in \mathcal{V}$, the objective is to determine the minimum CRF $c_x$ such that a re-encoded H.264 version of $v_i$ produces a bitrate not exceeding that of $s_i$, under the same target social video resolution $\rho_i^{\text{out}}$, frame rate, and codec, inferred directly from $s_i$ using \textit{ffmpeg}\footnote{\url{https://ffmpeg.org/}}. In contrast, the pixel format is set to YUV420p.

Formally, if we define the bitrate function $B: \mathcal{S} \rightarrow \mathbb{R}^{+}$, and let $E(v_i, \rho_i^{\text{out}}, c_x)$ denote the encoding operation on video $v_i$ with target resolution $\rho_i^{\text{out}}$ and CRF $c_x$, the estimated CRF $\hat{c}_i$ is computed as:

\begin{equation}
    \hat{c}_i = \min \left\{ c_x \in \mathbb{Z} \cap [c_{\text{min}}, c_{\text{max}}] \mid B\left(E(v_i, \rho_i^{\text{out}}, c_x)\right) \leq B(s_i) \right\}
    \label{eq1}
\end{equation}

where in \eqref{eq1}, the search of $\hat{c}_i$ is conducted in a linear sweep over $ c_x \in \mathbb{Z} \cap [c_{\text{min}}, c_{\text{max}}]$ with $c_\text{min}=21$ and $c_\text{max}=50$. This process is repeated for each of the $N$ videos. Each resulting triplet, composed by the $v_i$ resolution $\rho_i^{\text{in}}$, the $s_i$ resolution $\rho_i^{\text{out}}$, and the estimated CRF $\hat{c}_i$, is then stored in a dedicated database $\mathcal{D}$ defined as:

\begin{equation}
    \mathcal{D} = \{(\rho_i^{\text{in}}, \rho_i^{\text{out}}, \hat{c}_i)\}_{i=1}^{N}
    \label{eq2}
\end{equation}

\subsection{Social Network Encoding Emulation Module}

During the social network emulation stage, the framework produces compressed versions of a given set $\mathcal{U} = \{u_1, \dots, u_m\}$ of $M$ non-shared videos, where $\mathcal{V} \subset \mathcal{U}$ is possible. 

The emulation process utilizes the database $\mathcal{D}$, defined in \eqref{eq2}, to determine the appropriate output resolution $\rho_j^{*}$ and CRF parameter ${c}^*_j$ for each input video $u_j$, and emulate the target social video processing.

Let $\rho_j$ denote the video resolution of $u_j$. The framework first determines the target output resolution $\rho_j^*$ by searching within the database $\mathcal{D}$, for $i\in[1,N]$ as specified in Sect.~\ref{ssec:crf}, such that:

\begin{equation}
    \rho_j^* =
\begin{cases}
\rho_i^{\text{out}} & \text{if } \rho_j = \rho_i^{\text{in}} \text{ for } (\rho_i^{\text{in}}, \rho_i^{\text{out}}, \_) \in \mathcal{D} \\
\rho_k^{\text{out}}, & \text{where } k = \arg\min_i \|\rho_i^{\text{in}} - \rho_j\|_2 \quad \text{otherwise}
\end{cases}
\label{eq3}
\end{equation}

That is, if in \eqref{eq3}, an exact match $\rho_j= \rho_i^{\text{in}}$ exists in $\mathcal{D}$, $\rho_j^*=\rho_i^{\text{out}}$. Otherwise, the output resolution $\rho_i^{\text{out}}$, corresponding to the closest input resolution $\rho_i^{\text{in}}$ to $\rho_j$  (via Euclidean distance) is selected.

Next, since CRF values can vary for videos shared on social networks with the same resolution due to content differences (e.g., texture, flat regions, video duration), the CRF parameter ${c}^*_j$ is determined as the average of the CRF values in \(\mathcal{D}\) associated with the same output resolution $\rho_j^*$. This process is formalized in \eqref{eq4}.

\begin{equation}
    {c}^*_j = \frac{1}{|\mathcal{I}_{\rho_j^*}|} \sum_{l \in \mathcal{I}_{\rho_j^*}} \hat{c}_l,
\quad \text{where } \mathcal{I}_{\rho_j^*} = \{ l \mid \rho_l^{\text{out}} = \rho_j^* \}
\label{eq4}
\end{equation}

Finally, the social network-emulated video \(e_j \in \mathcal{E}\) is generated using $e_j = E(u_j, \rho_j^*, {c}_j^*)$.

During this phase, the encoding operator \( E(\cdot) \) uses the frame rate of the input video \( u_j \), the YUV420p pixel format, and H.264 encoding.

\section{Experimental Results}
\label{sec:Exp}

This section outlines the experimental setup and presents the results used to evaluate the proposed social network emulator's ability to replicate the video processing pipelines of major social media platforms.

All experiments are conducted using the FaceForensics++ (FF++) dataset~\cite{rossler_faceforensics_2019}, following the configuration introduced by Marcon et al.~\cite{marcon2021detection}. This configuration includes the original FF++ videos~\cite{rossler_faceforensics_2019} along with their corresponding versions after being shared on YouTube and Facebook. Specifically, the FF++ dataset used in~\cite{marcon2021detection} contains 1000 high-quality real videos, each manipulated using five face alteration techniques: Deepfake (DF)~\cite{deepfakes_faceswap_2018}, Face2Face (F2F)~\cite{thies2016face2face}, FaceSwap (FS)~\cite{Kowalski_FaceSwap_2018}, FaceShifter (FSH)~\cite{li2019faceshifter}, and NeuralTextures (NT)~\cite{thies2019deferred}. Among these, DF, FS, and FSH are identity-swapping techniques, while F2F and NT are based on facial reenactment.

The dataset~\cite{marcon2021detection} divides each real and fake video classes, into 720 videos for training, 140 for validation, and 140 for testing.
Each of the 280 validation and testing videos for the six classes (five fake and one real) were shared on both Facebook and YouTube, resulting in a total of 3360 shared videos.

Table~\ref{tab:nres} details the number of non-shared videos, per resolution, from the validation and test sets of FF++~\cite{marcon2021detection} that have been shared on Facebook and YouTube.

\begin{table*}[t]

\resizebox{\textwidth}{!}{%
\begin{tabular}{|c|c|c|c|c|c|c|c|c|c|c|c|c|c|c|c|c|}
\toprule
\textbf{Resolution} & 576$\times$480 & 600$\times$480 & 640$\times$480 & 642$\times$480 & 654$\times$480 & 656$\times$480 & 658$\times$480 & 704$\times$480 & 720$\times$480 & 832$\times$480 & 848$\times$480 & 854$\times$480 & 1280$\times$718 & 1280$\times$720 & 1920$\times$1080 & \textbf{Total} \\ \midrule
\textbf{\begin{tabular}[c]{@{}c@{}}Number of Validation\\ Videos\end{tabular}} & - & 6 & 39 & 1 & 1 & 7 & 1 & - & 3 & - & 1 & 14 & 1 & 54 & 12 & \textbf{140} \\ \midrule
\textbf{\begin{tabular}[c]{@{}c@{}}Number of Test\\ Videos\end{tabular}} & 15 & - & 50 & - & - & - & - & 5 & - & 13 & - & - & - & 35 & 22 & \textbf{140} \\ 
\bottomrule
\end{tabular}%
}
\caption{The distribution of FF++ \cite{marcon2021detection} validation and test-set video resolutions shared respectively on Facebook and YouTube.}
\label{tab:nres}
\end{table*}

Following the protocol described in~\cite{marcon2021detection}, we trained four CNNs: DenseNet~\cite{arulananth2024classification}, InceptionNet~\cite{szegedy2016rethinking}, XceptionNet~\cite{lu2022deep}, and ResNet-50~\cite{koonce2021resnet}; on the unshared training split of the FF++ dataset, training each model separately for each face manipulation technique. We then conducted two types of experiments.

The first experiment aimed to assess the performance degradation of deepfake detection models trained on unshared videos when tested on videos that had been shared on social media platforms. Subsequently, we evaluated the extent to which this performance could be recovered by fine-tuning the models on either shared videos or emulated videos generated using the proposed framework.

In the second experiment, we extended the evaluation to include videos shared on Facebook, YouTube, and BlueSky in 2025. Since the shared videos from~\cite{marcon2021detection} were uploaded to social media in 2021, this experiment aims to provide additional insight into the generalizability of the proposed emulation framework to future social network encoding pipelines.

During our experiments, we evaluated the performance degradation and recovery of the selected deepfake detectors using the True Positive Rate (TPR) on real videos and the True Negative Rate (TNR) on fake videos.

All hyperparameters were set according to the original paper~\cite{marcon2021detection}. Experiments were conducted on a server equipped with an NVIDIA GeForce RTX 3090 GPU, an Intel(R) Core(TM) i9-10940X 14-core CPU @ 3.30GHz, and 256~GB of RAM.

\subsection{Emulation of FaceForensics++ Videos Shared in 2021}
\label{ssec:exp1}

This section evaluates the ability of our framework to emulate the compression artifacts introduced by Facebook and YouTube on shared videos from the FF++ dataset~\cite{marcon2021detection}.

%Following the protocol described in~\cite{marcon2021detection}, we first trained four CNNs: DenseNet~\cite{arulananth2024classification}, InceptionNet~\cite{szegedy2016rethinking}, XceptionNet~\cite{lu2022deep}, and ResNet-50~\cite{koonce2021resnet}; to detect fake videos that had not been shared (NS) on social media platforms. 
We first trained four convolutional neural networks--DenseNet, InceptionNet, XceptionNet, and ResNet-50--to detect fake videos that had not been shared (NS) on social media platforms.
Each model was then fine-tuned using 280 validation videos (real and fake) from the corresponding fake classes in~\cite{marcon2021detection}, which had either been shared (S) on Facebook or YouTube, or processed by our emulation framework (EMU).

This process resulted in 60 distinct model configurations, covering every combination of network architecture (DenseNet, InceptionNet, XceptionNet, and ResNet-50), fake video class from FF++ (DF, F2F, FSH, FS, and NT), and compression type (NS, S, or EMU). We evaluated each model configuration on the corresponding fake video class from the FF++ test set, using videos shared via Facebook and YouTube.

\begin{figure*}[t]
    \centering
    \begin{subfigure}[b]{0.89\textwidth}
        \centering
        \includegraphics[width=\linewidth]{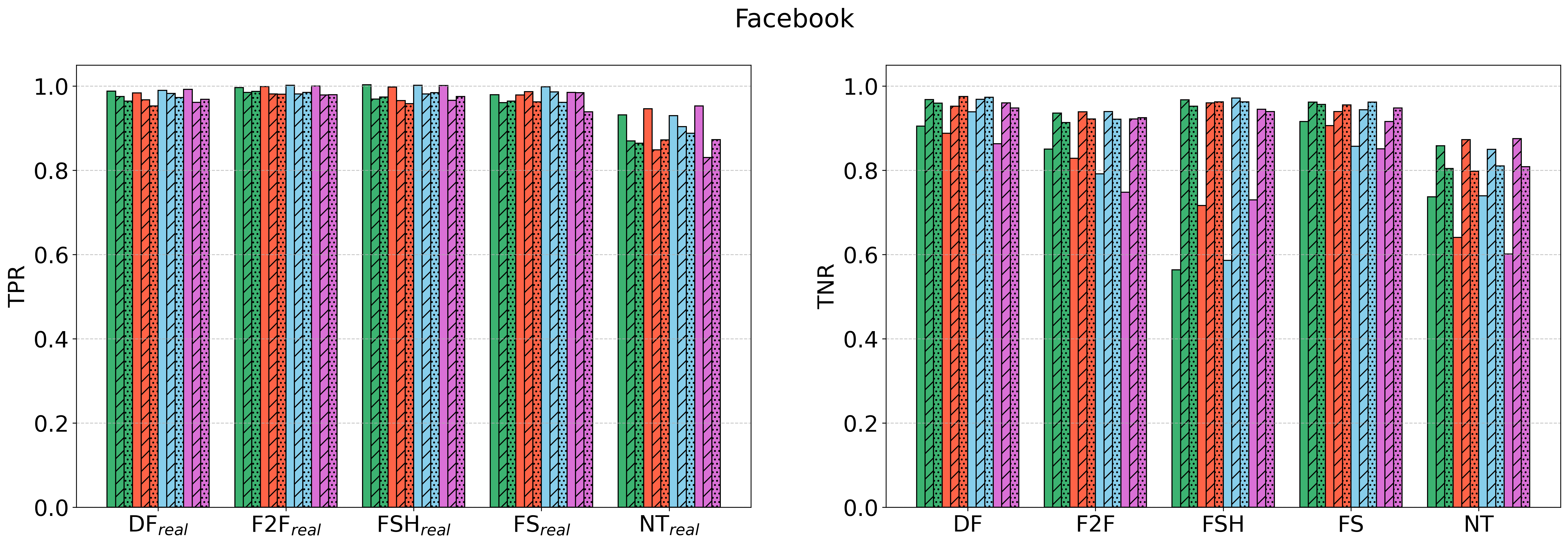}
        \caption{}
        \label{fig:exp1_fb}
    \end{subfigure}
    \vfill
    \begin{subfigure}[b]{0.89\textwidth}
        \centering
        \includegraphics[width=\linewidth]{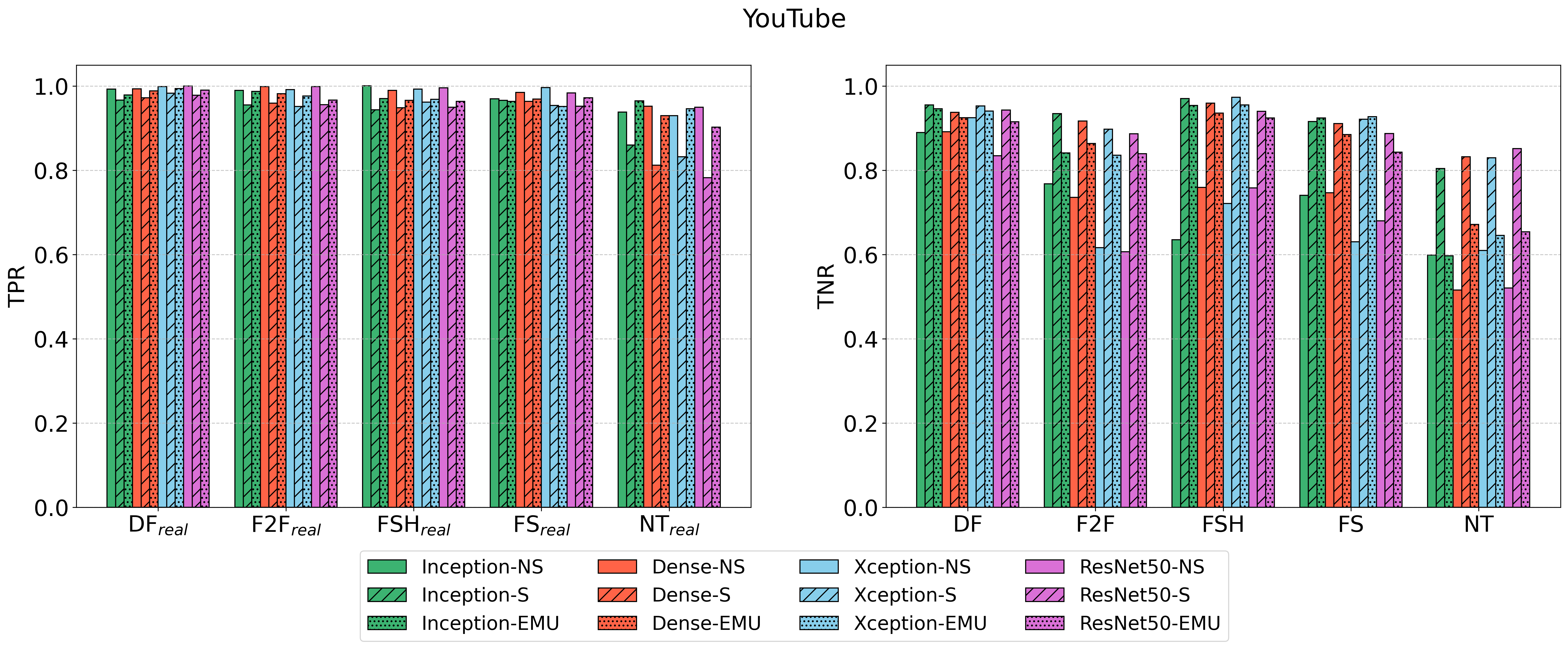}
        \caption{}
        \label{fig:exp1_yt}
    \end{subfigure}
    \caption{TPR and TNR obtained by four deepfake detectors on videos shared on Facebook and YouTube. Models were trained under three conditions: (NS) trained on videos not shared on social media; (S) fine-tuned on videos shared on social media; and (EMU) fine-tuned on videos with emulated social media sharing. Real videos corresponding to the tested fake video class, are denoted with the subscript ${real}$.}
    \label{fig:exp1}
\end{figure*}

Figure~\ref{fig:exp1} presents the True Positive Rate (TPR) and True Negative Rate (TNR) results for the FF++ test set, consisting of both real and fake videos shared on Facebook and YouTube. TPR values were computed on real videos associated with each corresponding fake video class (DF, F2F, FSH, FS, and NT), and are denoted using the subscript $_{\text{real}}$.

On real videos shared on Facebook (Fig.~\ref{fig:exp1_fb}), all models achieved similar TPR, regardless of the training data used. In contrast, the TNR on fake videos was significantly reduced when models were trained exclusively on non-shared (NS) videos. However, fine-tuning on either shared (S) or emulated (EMU) videos led to substantial and comparable improvements in TNR performance.

Consistent patterns were observed for videos shared on YouTube (Fig.~\ref{fig:exp1_yt}). Nonetheless, in the case of the NeuralTextures (NT) manipulation, the TNR gains obtained from models fine-tuned on emulated (EMU) videos were consistently lower than those achieved by fine-tuning on actually shared (S) videos. Conversely, for the real videos in the NT test set, models fine-tuned on emulated data achieved higher TPR. We hypothesize that this discrepancy may be due to sub-optimal thresholding, as a fixed classification threshold of 0.5 was applied across all detectors, in line with the original study~\cite{marcon2021detection}.

These results suggest that our framework can serve as a practical alternative to direct platform-based data collection, which is often constrained by API limitations and associated operational costs. As shown in Fig.~\ref{fig:exp1}, our method can effectively estimate the parameters required for realistic video emulation. Thus, the resulting emulated videos can be used to fine-tune deepfake detectors, improving their ability to identify manipulated content even after it has been shared on social media platforms.

\subsection{Emulation of Videos Shared on Social Networks in 2025}
\label{ssec:exp2}

\begin{figure*}[t]
\centering
\begin{subfigure}[b]{0.89\textwidth}
\centering
\includegraphics[width=\linewidth]{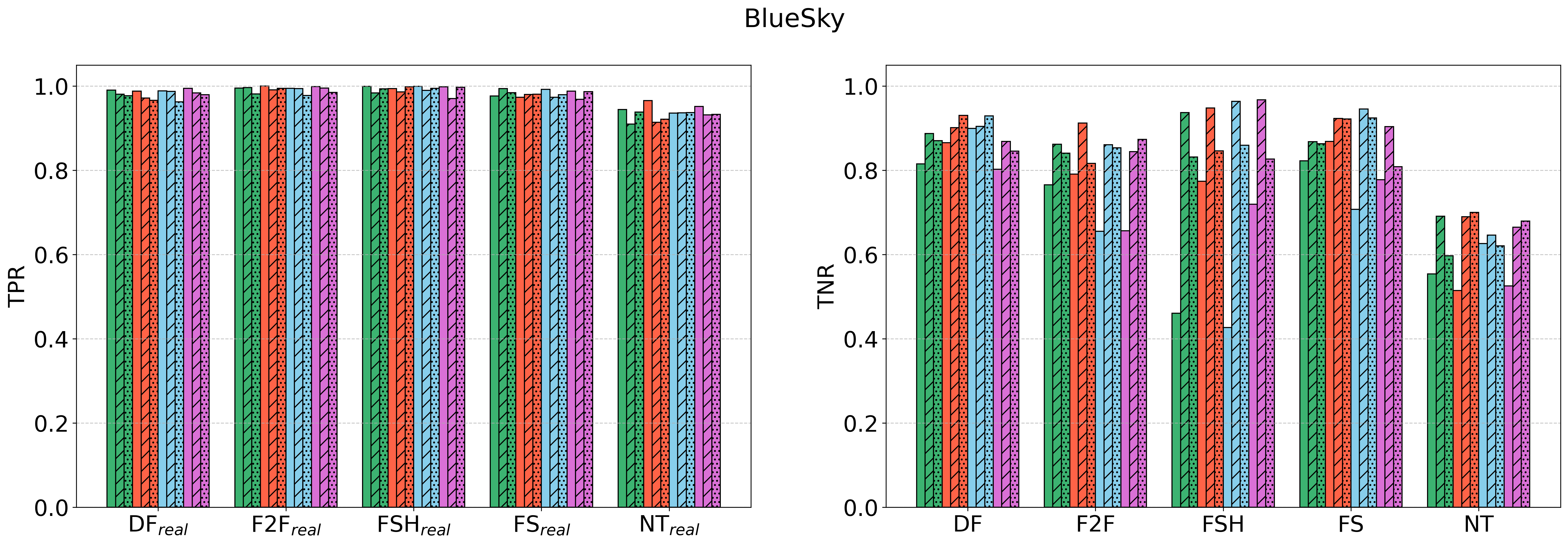}
\caption{}
\label{fig:exp2_bs}
\end{subfigure}
\vfill
\begin{subfigure}[b]{0.89\textwidth}
\centering
\includegraphics[width=\linewidth]{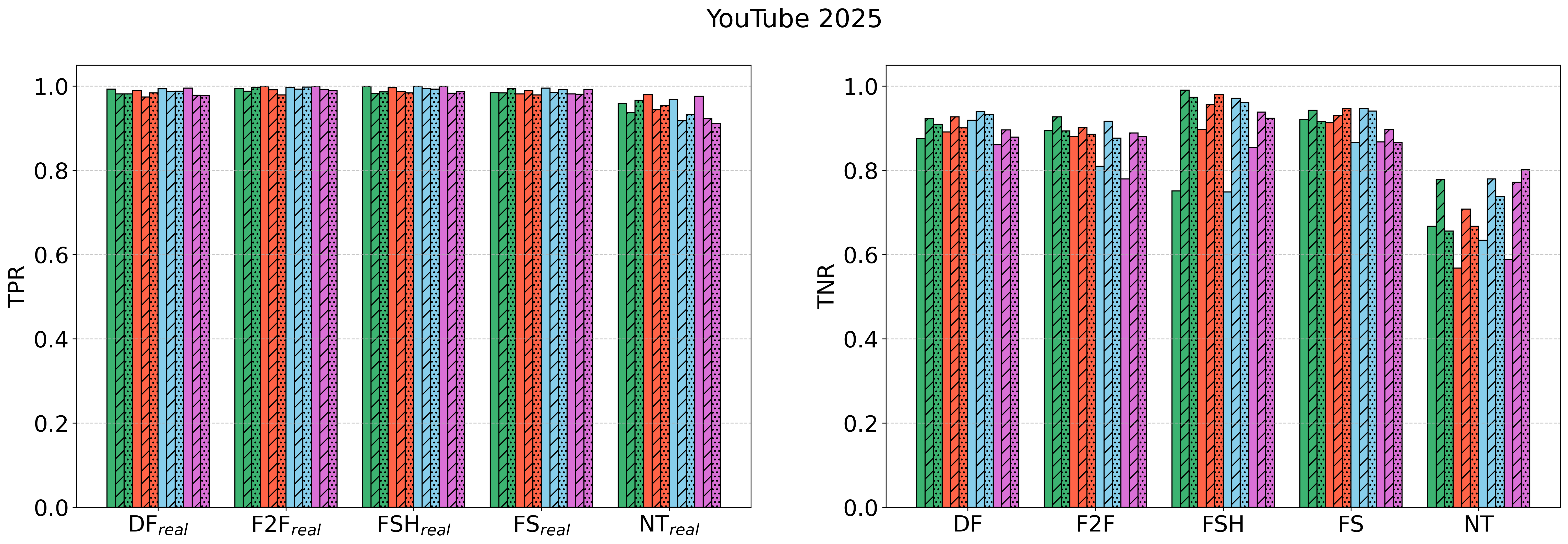}
\caption{}
\label{fig:exp2_yt}
\end{subfigure}
\vfill
\begin{subfigure}[b]{0.89\textwidth}
\centering
\includegraphics[width=\linewidth]{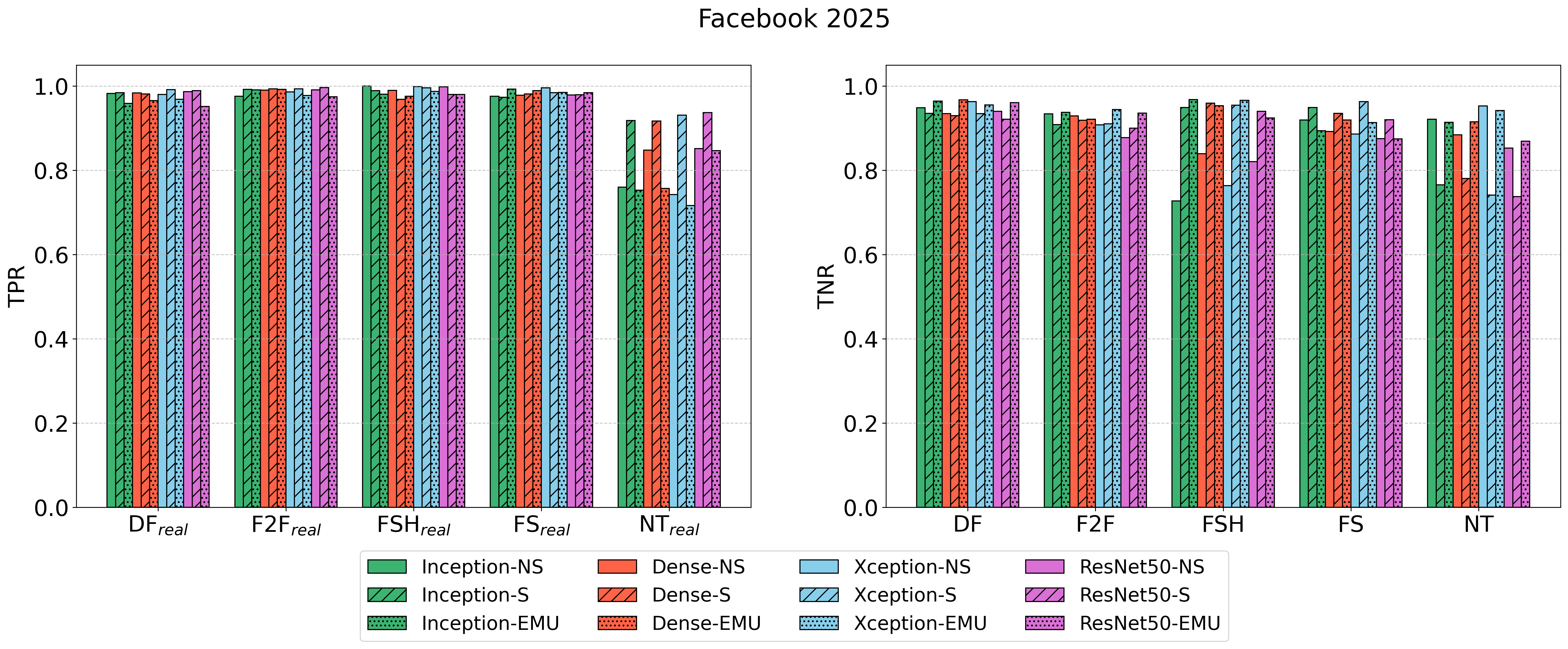}
\caption{}
\label{fig:exp2_fb}
\end{subfigure}
\caption{TPR and TNR obtained by detectors evaluated on videos shared, during 2025, on BlueSky, YouTube, and Facebook. Training conditions include: (NS) trained on non-shared videos; (S) fine-tuned on videos shared on the respective platform; and (EMU) fine-tuned on videos created with our emulation framework. Real videos corresponding to the tested fake video class, are denoted with the subscript ${real}$.}
\label{fig:exp2}
\end{figure*}

Since the original FF++ videos~\cite{marcon2021detection} were shared using the 2021 versions of Facebook and YouTube, we conducted additional experiments to evaluate our framework’s ability to adapt to the current (2025) versions of these platforms, as well as to other social networks. To this end, we repeated the previous set of experiments using videos that we shared on Facebook, YouTube, and BlueSky in 2025. The results of these experiments are presented in Fig.~\ref{fig:exp2}.

Overall, the results presented in Fig.~\ref{fig:exp2} are consistent with those reported in Section~\ref{ssec:exp1}. This consistency is particularly evident for videos shared on BlueSky (Fig.~\ref{fig:exp2_bs}), as we did not encounter any API quota limitations on this platform and were therefore able to share the entire validation set for all real and fake video classes. However, this was not the case for YouTube (Fig.~\ref{fig:exp2_yt}) and Facebook (Fig.~\ref{fig:exp2_fb}), where API constraints limited us to sharing only 50 of the 140 validation videos for each FF++ fake and real class.

\begin{table*}[t]
\resizebox{\textwidth}{!}{%
\begin{tabular}{|cl|c|c|c|c|c|c|c|c|c|c|c|c|c|c|c|c|}
\toprule
\multicolumn{2}{|c|}{\textbf{Resolution}} & 576$\times$480 & 600$\times$480 & 640$\times$480 & 642$\times$480 & 654$\times$480 & 656$\times$480 & 658$\times$480 & 704$\times$480 & 720$\times$480 & 832$\times$480 & 848$\times$480 & 854$\times$480 & 1280$\times$718 & 1280$\times$720 & 1920$\times$1080 & Total \\ \midrule
\multirow{2}{*}{\textbf{BlueSky}} & val & - & 6 & 39 & 1 & 1 & 7 & 1 & - & 3 & - & 1 & 14 & 1 & 54 & 12 & \textbf{140} \\ \cmidrule{2-18} 
 & test & 15 & - & 50 & - & - & - & - & 5 & - & 13 & - & - & - & 35 & 22 & \textbf{140} \\ \midrule
\multirow{2}{*}{\textbf{Facebook 2025}} & val & - & - & 26 & - & - & - & - & 2 & - & - & - & - & - & 25 & 7 & \textbf{50} \\ \cmidrule{2-18} 
 & test & 7 & - & 28 & - & - & - & - & 3 & - & 7 & - & - & - & 12 & 13 & \textbf{50} \\ \midrule
\multirow{2}{*}{\textbf{YouTube 2025}} & val & 4 & - & 18 & - & - & - & - & 1 & - & - & - & - & - & 20 & 4 & \textbf{50} \\ \cmidrule{2-18} 
 & test & 4 & - & 20 & - & - & - & - & 2 & - & 7 & - & - & - & 8 & 9 & \textbf{50} \\ \bottomrule
\end{tabular}%
}
\caption{The distribution of FF++ \cite{marcon2021detection} validation (val) and test-set video resolutions shared respectively on BlueSky, Facebook, and YouTube during 2025.}
\label{tab:nres2}
\end{table*}

By analyzing the resolution distributions of FF++ validation videos~\cite{marcon2021detection} shared on BlueSky, YouTube, and Facebook (see Table~\ref{tab:nres2}), we observed that the inconsistent performance on Facebook and YouTube, visible in Fig.~\ref{fig:exp2}, correlates with the number of validation videos used to estimate the CRF parameter $\hat{c}$ at specific output resolutions $\rho^{\text{out}}$, such as {1280$\times$720} and {640$\times$480}. Specifically, for Facebook and YouTube, $\hat{c}$ was estimated using at most 26 videos, nearly half the number available for BlueSky, as reported in Table~\ref{tab:nres2}.

Moreover, when comparing the resolution distributions of the validation and test sets (Table~\ref{tab:nres2}), we found that the most frequent resolutions in the test set, including {1280$\times$720} and {640$\times$480}, are also those with the greatest discrepancies in the number of videos used for CRF estimation across BlueSky, Facebook, and YouTube.

This is especially relevant because, while the resolution transformation applied by a social network tends to be consistent, the CRF value used during recompression by social networks can vary substantially depending on content characteristics such as texture complexity, the presence of flat regions, and video duration. Consequently, when the number of available videos per resolution is limited, the resulting CRF estimate $c^*$ may be suboptimal for accurately emulating the target platform’s compression.

Motivated by these findings, and to provide practical guidance for using our emulator effectively, we present an ablation study in Section~\ref{ssec:ablation} to determine the minimum number of videos per resolution required for reliable CRF estimation.

\subsection{Ablation Study}
\label{ssec:ablation}
\begin{figure*}[t]
    \centering
    \begin{subfigure}[b]{0.4\textwidth}
        \centering
        \includegraphics[width=\linewidth]{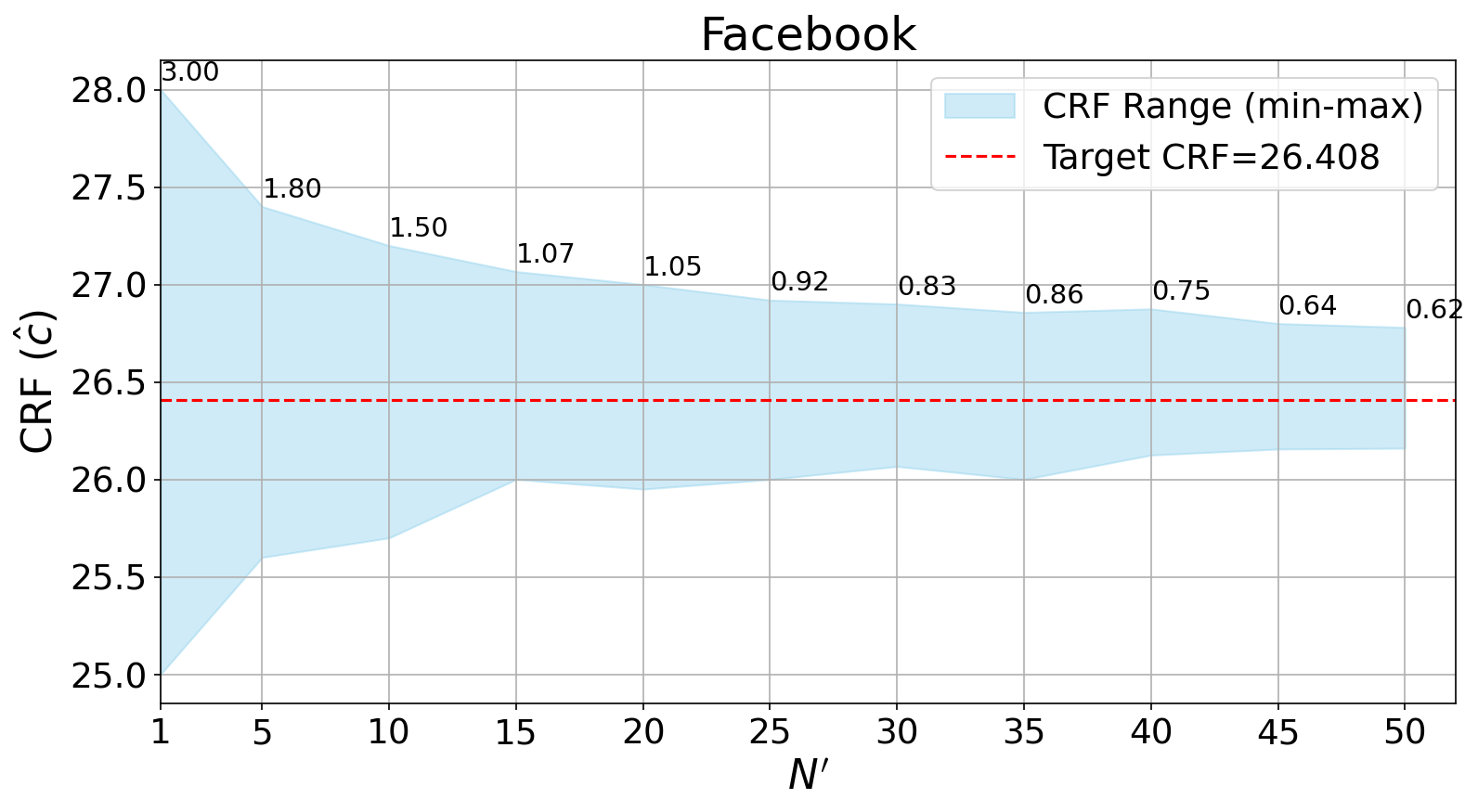}
        \caption{}
        \label{fig:exp5_fb}
    \end{subfigure}
    \hfill
    \begin{subfigure}[b]{0.4\textwidth}
        \centering
        \includegraphics[width=\linewidth]{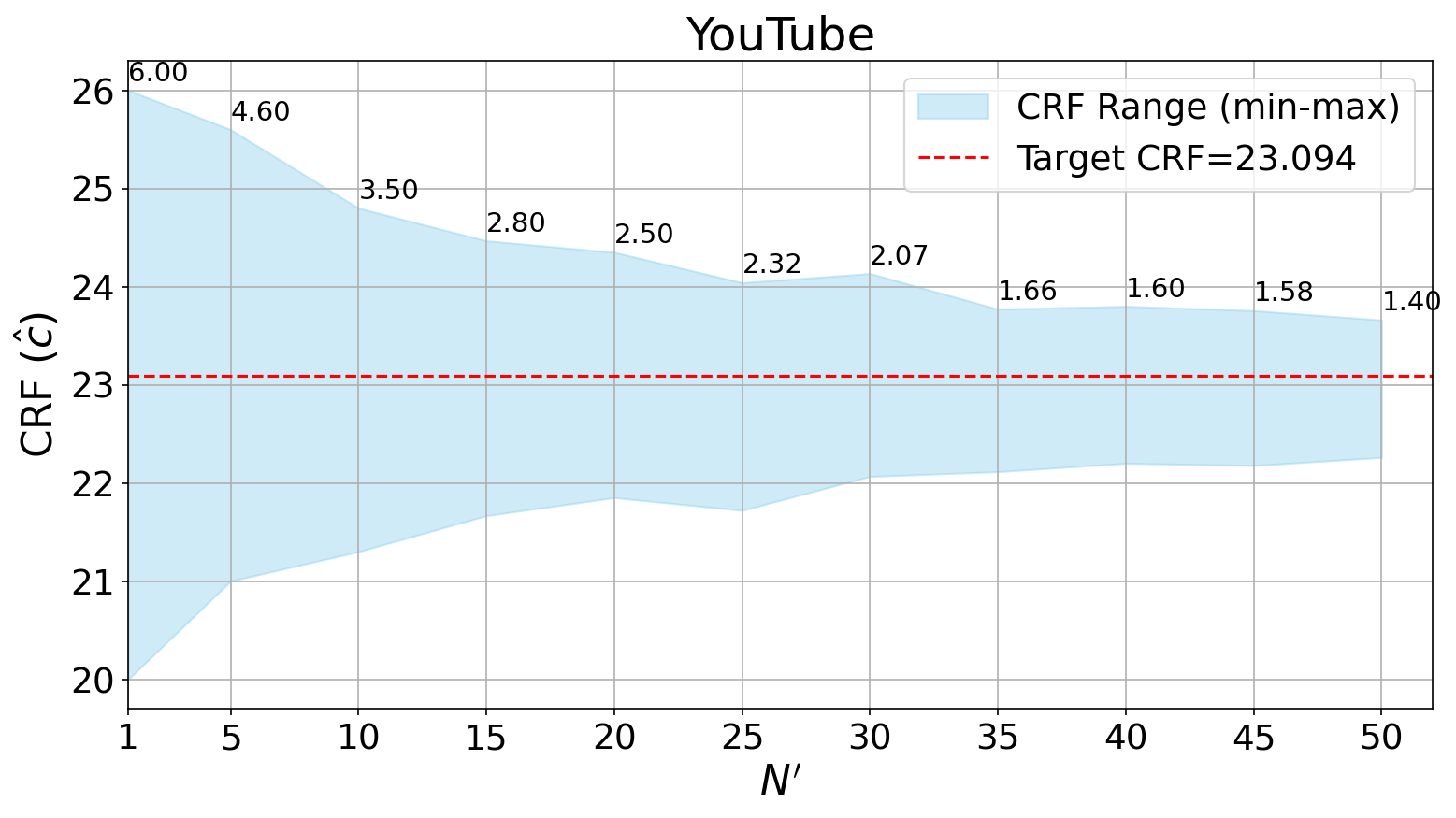}
        \caption{}
        \label{fig:exp5_yt}
    \end{subfigure}
    \caption{Stability analysis of CRF parameter estimation across bootstrap validation, using shared videos from FF++ \cite{marcon2021detection}. The plot displays the distribution of estimated CRF parameters $\hat{c}$ over 1000 bootstrap iterations as a function of sample size ($N'$). The shaded blue area represents the CRF range, with its explicit value reported at the top of it. The values stabilize and exhibit lower variance for $N' \geq 30$.}
    \label{fig:exp5}
\end{figure*}%

To evaluate how the number of shared videos per resolution affects the CRF estimation, we conducted a bootstrap \cite{efron1992bootstrap} ablation study. In this study, we assessed the stability of the estimated CRF parameter $c^*$ by repeatedly sampling subsets of $N' \in [1, 50]$ videos. For each value of $N'$, we performed 1000 iterations, randomly selecting a new subset of $N'$ videos, and estimating the CRF parameter $\hat{c}$ using \eqref{eq1}. Thus, we computed the range of the estimated CRF values $\hat{c}$ for each subset size $N'$. 

We conducted this experiments on videos with resolution $1280 \times 720$ from the FF++ validation split, as this resolution provided the largest pool of shared videos, 54 in total.

Figure~\ref{fig:exp5} illustrates the convergence behavior of the estimated CRF values $\hat{c}$ as the sample size $N'$ increases, using Facebook and YouTube videos from~\cite{marcon2021detection}. We observed that for $N' \geq 30$, the CRF estimates stabilized within a consistent range and exhibited significantly reduced variance. 
% This result suggests that a minimum of 30 shared videos per resolution is necessary to obtain a reliable approximation of a platform’s video processing pipeline.

We therefore conclude that, for our framework to provide a practical and scalable emulation of social network video compression, at least 30 shared videos per resolution are required. This conclusion is also consistent with the observations made in Section~\ref{ssec:exp2}.

\section{Conclusions and Future Works}
\label{sec:Conc}
In this work, we introduced a novel framework designed to bridge the gap between laboratory settings and real-world multimedia forensics applications, specifically focusing on the detection of deepfake videos shared on social networks. Our approach addresses the significant challenges associated with direct data collection from social media platforms (such as API limitations, high costs, and rate-limiting) by accurately emulating platform-specific video compression and resizing pipelines.

We demonstrated that, by analyzing fewer than 50 shared videos per resolution, our framework can reliably estimate the encoding parameters required to replicate a given platform's video processing. These parameters are then used to process large-scale datasets locally, generating realistically degraded videos for model fine-tuning and evaluation. This eliminates the need for extensive direct sharing and enables scalable experimentation.

Our experiments, conducted on the widely-used FaceForensics++ dataset~\cite{marcon2021detection}, validate the effectiveness of the proposed approach. We showed that deepfake detectors fine-tuned on videos processed by our emulator achieve performance comparable to those fine-tuned on actually shared videos from platforms such as Facebook and YouTube. This finding holds for both legacy data from 2021~\cite{marcon2021detection} and videos shared in 2025 on updated versions of Facebook and YouTube, as well as on emerging platforms such as BlueSky. Additionally, our analysis provides practical insights into the minimum number of shared videos required per resolution to ensure stable and reliable CRF estimation.
The developed code will be released as an open-source, user-friendly Python library to support broader adoption within the research community.

Future work will focus on expanding and maintaining the proposed framework, with particular emphasis on enabling the estimation of additional encoding parameters, such as frame rate and pixel format. Additionally, we plan to broaden our evaluation to encompass a wider range of platforms, including other major social networks (e.g., TikTok and X, formerly Twitter) as well as popular messaging applications like Telegram, WhatsApp, and Slack.

\begin{acks}
{This work was partially supported by the European Union under the Italian National Recovery and Resilience Plan (NRRP) of NextGenerationEU (PE00000014 - program “SERICS”), and the NGI Sargasso project DeepShield (101092887). }
\end{acks}

%\newpage
\bibliographystyle{ieeetr}
\balance
\bibliography{reference}
\end{document}